\documentclass[letterpaper, 10 pt, conference]{ieeeconf}  
\IEEEoverridecommandlockouts                               
\usepackage{graphicx}                       
\usepackage{graphics}                       
\usepackage{epsfig}                         
\usepackage[tight,footnotesize]{subfigure}  
\usepackage{float}
\usepackage{amssymb,amsmath}
\usepackage{mdwmath}
\usepackage{commath}                        
\usepackage{eqparbox}
\usepackage{amsmath,amsfonts,amssymb}

\newcommand{\ie}{{\em i.e.\ }}

\newcommand{\et}{{\em et al.\ }}

\newcommand{\beq}{\begin{equation}}
\newcommand{\eeq}{\end{equation}}
\newcommand{\bear}{\begin{eqnarray}}
\newcommand{\bears}{\begin{eqnarray*}}
\newcommand{\eear}{\end{eqnarray}}
\newcommand{\eears}{\end{eqnarray*}}
\newcommand{\bdm}{\begin{displaymath}}
\newcommand{\edm}{\end{displaymath}}
\newcommand{\lba}{\left[\begin{array}}
\newcommand{\ear}{\end{array}\right]}

\usepackage{bm}   
\usepackage{siunitx}
\usepackage{stfloats}                       
\usepackage{hyperref}
\usepackage{cite}                           
\usepackage[symbol]{footmisc}
\usepackage[T1]{fontenc} 
\usepackage{tabularx}
\usepackage[table]{xcolor}
\usepackage{longtable}
\usepackage{booktabs}       					
\usepackage{xcolor,colortbl}               		

\title{\LARGE \bf Dynamic Interaction Probabilistic Movement Primitives }
\author{Shuangda Duan, Longxin Chen, Hongmin Wu, Yaxiang Wang, Xuan Zhao, and Juan Rojas. \\
}
\begin{document}
\maketitle
\thispagestyle{empty}
\pagestyle{empty}
\bstctlcite{IEEEexample:BSTcontrol}
\begin{abstract}
Human-robot collaboration is on the rise. Robots need to increasingly improve the efficiency and smoothness with which they assist humans by properly anticipating a human's intention. To do so, prediction models need to increase their accuracy and responsiveness. This work builds on top of Interaction Movement Primitives with phase estimation and re-formulates the framework to use dynamic human-motion observations which constantly update anticipatory motions. The original framework only considers a single fixed-duration static human observation which is used to perform only one anticipatory motion. Dynamic observations, with built-in phase estimation, yield a series of updated robot motion distributions. Co-activation is performed between the existing and newest most probably robot motion distribution. This results in smooth anticipatory robot motions that are highly accurate and with enhanced responsiveness. 
\end{abstract}
\section{Introduction}\label{sec:introduction}
In human-robot interaction (HRI) and human-robot collaboration (HRC) anticipation is key for smoother collaboration. Anticipatory motions are a result of predictions a robot makes as it observes its human counter-part. However, a main drawback in predicting is the trade-off between achieved trajectory and goal-pose accuracy and longer observation periods. The sooner a human trajectory is predicted, the sooner a robot can offer assistance. However, quick predictions may lead to inaccurate robot motions; whilst longer observations may achieve higher accuracy but with lags in responsiveness. Most of the work in HRI anticipation uses fixed-time human observation windows. In this work, we introduce dynamic human observation window as a general-case scenarios for agent observations (static windows, on the other hand, are just a special case). 
In particular, we are interested to resolve situations when a multiple human motions begin with a similar motion but divert later on. This is typical in situations where the human moves initially to grasp or deliver an object, but is waiting to better understand environmental conditions before transitioning to a final motion. This may occur when external factors like others humans are also working in the current environment (see \cite{2017AutBot-Luo-UnsupEarlyPred_HRICollab}) for example.
\begin{figure}[t]
  	\centering
		\includegraphics[width=\linewidth]{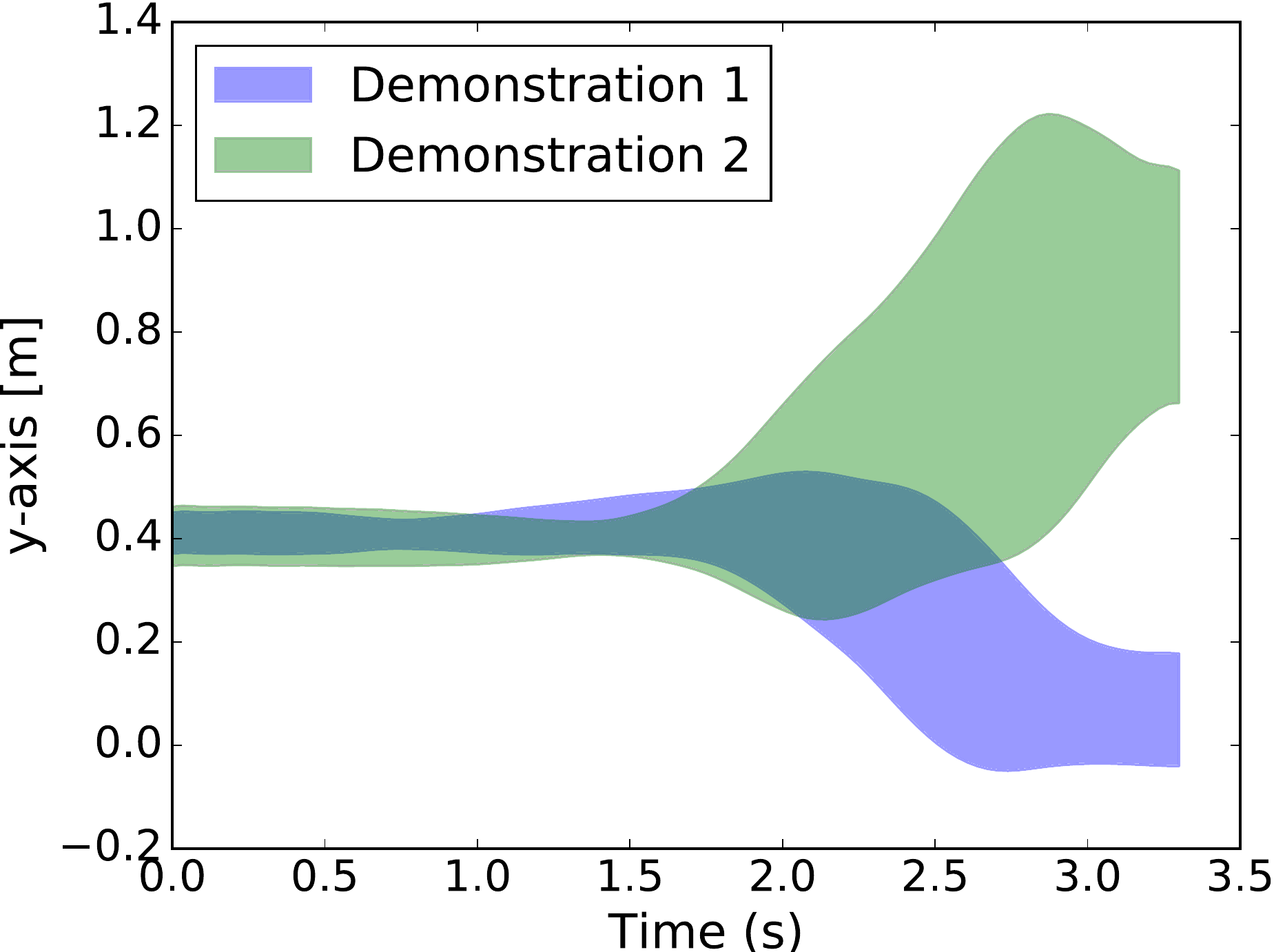}
    	\caption{Example of two human motion trajectories that start with a similar path but diverge in the latter half. Static fixed-duration observation techniques do not identify changes that occur in the post-observation period.}
    	\label{fig:problem_state}
\end{figure}
In HRI and HRC, numerous approaches have been used to generate robot motion in response to human motion observations. The research is characterized by estimating the human state to generate an appropriate and corresponding robot motion. Some predictive works have focused on optimizing safety and avoiding predicted human-occupancy spaces to avoid collisions \cite{2019IJRR-Park-IntAware_MotPlan_HumMotPred,2017AutBot-Luo-UnsupEarlyPred_HRICollab,2013IROS-Mainprice-HRI_EarlyPredHumMotn}. Other works like Interaction Probabilistic Motion Primitives (IProMPs) \cite{2014Humanoids-Maeda-LearnIP,2017IJRR-Maeda-PhaseEstimation,2015ICRA-ewerton-LearnMultCollabTasks_MixtureInteractionPrimitives} generate predictive robot motions based on one-time fixed observations of human motions to achieve physical interactions like: handover tasks. Interactive meshes (IMs)\cite{2017ICRA-Vogt-SysLearnCont_HRI_HHDems} learn an interaction model from human-to-human demonstrations that can be used to continuously update a robot's motion during collaborative tasks. IMs model complex interactions; however they lack (timely) responsiveness.
In \cite{2013NIPS-Paraschos_Peters-ProMPs}, Parascho's \et introduced the Probabilistic Movement Primitives (ProMPs) framework to compose complex robot skills from basic movements in a modular control architecture. The framework provided a single unified formulation that enabled probabilistic operations to support the parallel and smooth co-activation of MPs. Tasks could then be generated as sequences of simple or simultaneously activated skills. The ProMP formulation, like Dynamic Motion Primitives \cite{2017IJRR-Maeda-PhaseEstimation} can do temporal and velocity modulation; however unlike ProMPs, DMPs do not address the inverse problem of estimating the phase itself. Basis functions encode positions which are critical for the tractability of interaction primitives since estimating the forcing function of a human (a DMP requirement) is non trivial. 
IProMPs where initially presented in \cite{2014Humanoids-Maeda-LearnIP}. The goal of the framework is to enable a robot assistant to adapt and learn new interactive skills on demand. Imitation learning is leveraged in collaboration tasks and ProMPs are used to generate distributions from human motion observations. The distributions serve as a prior model in a lower dimensional weight space. The model is used to recognize the intended motion of the human agent and to generate a movement primitive for the robot. By using the IProMP framework, trajectories from the human and the robot can be naturally correlated thus simplifying the framework's complexity.
Maeda \et extended the IProMP framework to track human movement progress \cite{2017IJRR-Maeda-PhaseEstimation}. A temporal rescaling factor is estimated through human observations to improve the robot motion regression. The observation however is static. Only a single human-motion observation is used to regress the corresponding robot motion. The system is unable to dynamically adapt its prediction given new observations. 
More recently in \cite{2018RAL-Manschitz-MixtureAttractors}, Manschitz \et enhanced the DMP formulation so as to adapt to different objects. Attractors can be represented in different coordinate frames. They can also continuously activate a set of attractors over time by solving a convex optimization problem. The co-articulation of movements is also supported explicitly. The work however assumes distributions for the entire task which are both time-aligned and temporally-modulated. In contrast our robot's adaption to human motions leverages dynamic observation windows (DOWs) that yield sets of possible robot motion distributions. The most likely new robot motion distribution is merged with the existing motion through blending \cite{2013NIPS-Paraschos_Peters-ProMPs} to yield continuous, more accurate and more responsive robot motion updates. 
\begin{figure}[t]
  	\centering
		\includegraphics[width=\linewidth]{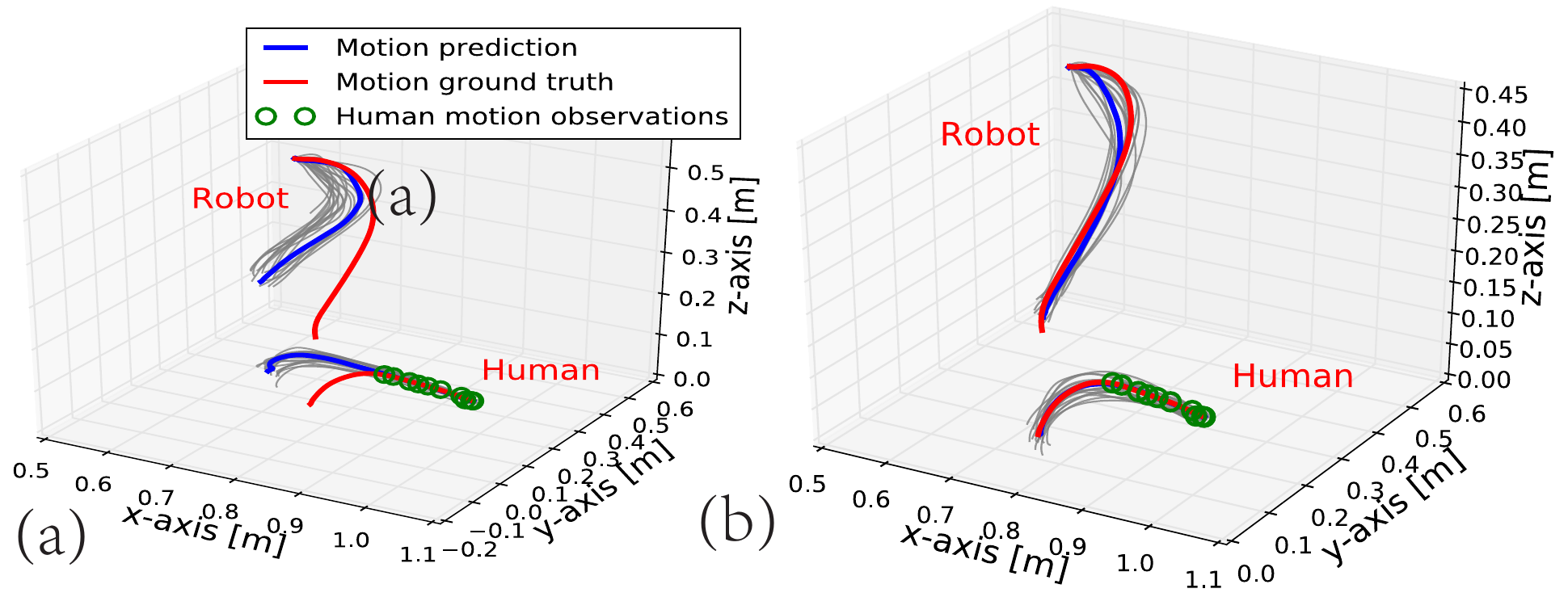}
    	\caption{Two tasks start with a similar distribution but vary in the latter part of the trajectory. Left: a wrongly anticipated robot motion given a single fixed-duration human observation. Right: correct prediction.}
    	\label{fig:demo_3d_compare}
\end{figure}    
\begin{figure}[b]
  	\centering
		\includegraphics[width=8cm, height=5cm]{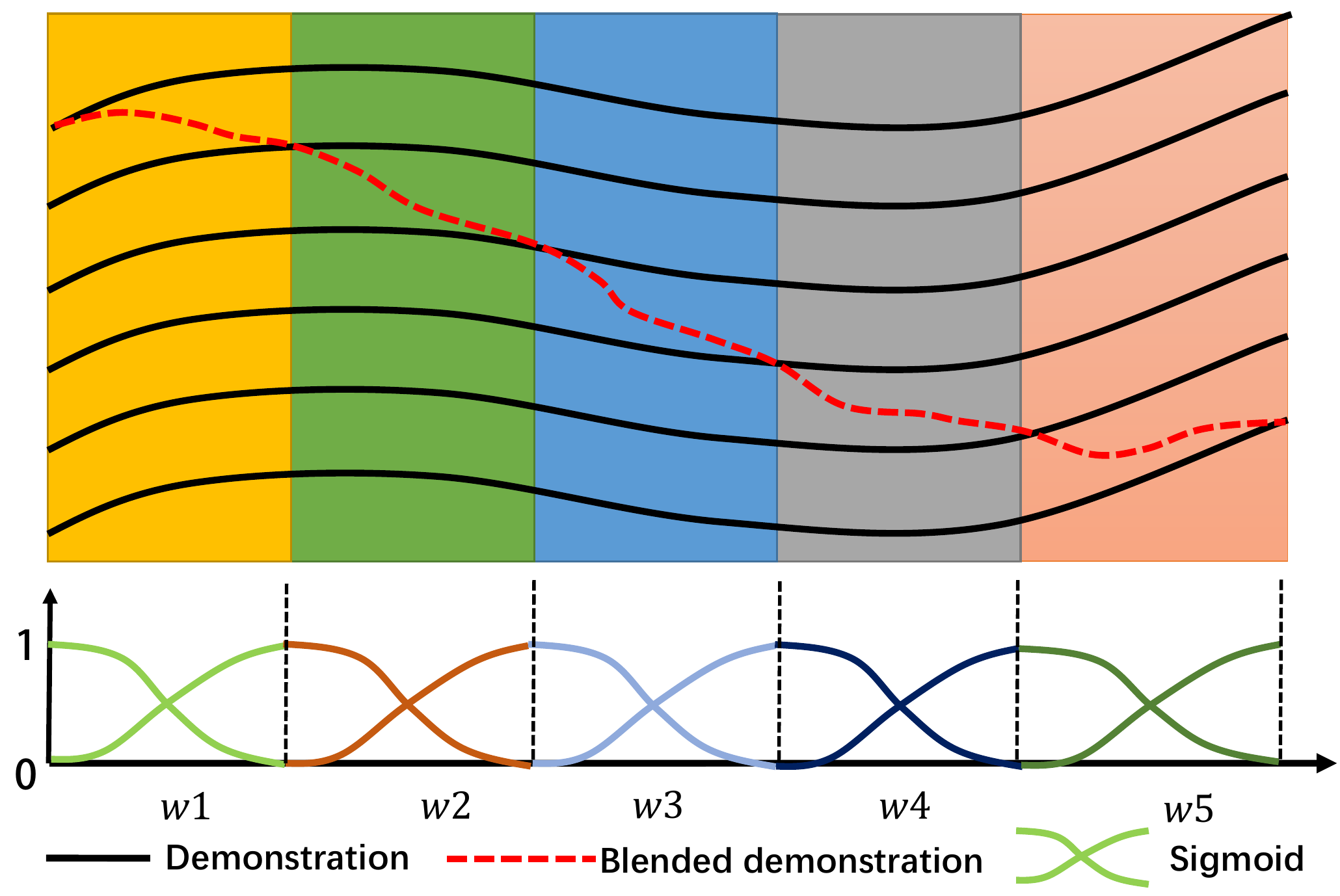}
    	\caption{DOW yield six robot distributions. From top-to-bottom, each distribution's probability gradually increases. Iterative blending is performed by co-activating the current distribution with the incoming distribution as presented in Eqtn. \ref{eqtn:prod_dists_activation}.
    	}
    	\label{fig:blending}
\end{figure}    
In this work we study the effectiveness of dynamic observations for human motions in HRI tasks.
Our contribution extends Interactive Primitives via the Probabilistic Movement Primitives to use a dynamic time-windows for human observations. Compared with the single fixed-time (static) formulation, our dynamic approach yields anticipatory robot motions with higher accuracies (throughout the trajectory and at the end-pose), temporal responsiveness, and good phase estimation in the robot's motion. We also contribute a metric that aids in the selection of a time-window duration that minimizes positioning and phase estimation errors. Though the re-formulation for dynamic situations is simple, it results in superior responsiveness and accuracy achievements for the robot's anticipatory motions, especially when human motions share similar initial motions but divert later in the task.  

We use the IProMP framework with phase estimation to model MPs via Gaussian distributions. Human observations are captured through a dynamic time-window whose duration is determined by a parameter (Sec. \ref{subsec:IProMP_Ndim}). Phase estimation is limited to the dynamic window duration (Sec. \ref{subsec:phase_est_w_dw}). Once the human observation is updated based on the human phase, we perform the robot task recognition by selecting the posterior distribution with the highest probability (Sec. \ref{subsec:task_recognition}). Blending is then performed iteratively for each new dynamic observation (Sec. \ref{sec:trajectory_blending}). To blend, we co-activate the current distribution with the incoming distribution by computing the product of the distributions as a function of a smooth activation function. Fig. \ref{fig:blending} shows how five co-activations occur in a task. The output is a smooth transition for the iteratively updated robot trajectory. 

In our experiments (Sec. \ref{sec:experiments}) we compare the performance of the dynamic time-window IProMP formulation with that of a static time-window formulation across two experiments: (a) tasks with a uniform trajectory pattern and (b) tasks where trajectories are similar at the beginning of the task but diverge later in the task. For each experiment, we compare accuracy gain as a function of dynamic window-time duration for three measurements: (i) robot joint angle configuration at each time-step, (ii) Cartesian position at the final goal position, and (iii) phase estimation error. Finally, a weighted sum error metric is used to select the most optimal dynamic window time duration across all metrics. 
In the first experiment, we tests hand-over tasks for three different objects. We vary the duration of the dynamic time-window to study its effect on robot position accuracy as well as phase estimation error. 
We found that for DOWs, we achieve average accuracies that would only be possible if the static window duration were 90\% or more of the human motion. 
In the second experiment, we test the responsiveness of our system by testing on sets of trajectories that have common paths at the beginning but deviate later on.
We found that DOWs yields a much more responsive system; one that is able to adapt correctly in situations where multiple learned human distributions initiate with similar paths but deviate later in the trajectory. 
Our work thus shows that more accurate and responsive motions can be attained in probabilistic formulations for HRI/HRC by dynamically updating the human observations. 

Our paper is organized as follows: in Sec. \ref{sec:ipromp}, we introduce the Probabilistic Motion Primitives framework and its extensions to Interaction Primitives and Phase Estimation. In Sec. \ref{sec:trajectory_blending}, we explain how blending can be done for dynamic observation settings. Sec. \ref{sec:experiments} presents two experiments and associated results. Finally, in Sec. \ref{sec:conclusion} we highlight key lessons learned in this work.
\section{Interaction Motion Primitives with Phase Estimation and Dynamic Observations}\label{sec:ipromp}
In HRC tasks, IProMPs generate a robot collaborative motion based on the prediction from a set of partial human motion observations. In this section, we introduce IProMPs by first explaining probabilistic movement primitives for a single dimension. We then expand the case for multiple dimensions and also describe the formulation for interactive scenarios. Later we incorporate phase estimation with dynamic time-window human observations and conclude by explaining parameter learning. 
\subsection{Probabilistic Movement Primitives for a Single Dimension}\label{subsec:IProMP_1dim}
ProMPs summarize patterns across demonstrations in a probabilistic manner. ProMPs capture correlations across data dimensions leading to a probability distribution over trajectories. Representing variance information correctly is critical as it reflects variations in movement execution across time steps. For each time step, a single dimension position is represented by $y_t \in \mathbb{R}^1$ and a trajectory of $T$ time steps as $\bm{y}_{1:T}$. A parameterization of $\bm{y}_{1:T}$ in a lower dimensional weight space is given as as a linear basis function model of $n$ Gaussian basis functions and weights $\bm{\omega}$ according to:
\begin{equation}
  \begin{aligned}
  y_t &= \bm{\psi}_t^T \bm{\omega} + \epsilon_y,\\
  p( \bm{y}_{1:T} | \bm{\omega}) &= \prod\limits_1^T \mathcal{N}(y_t | \bm{\psi}_t^T \bm{\omega}, \sigma_y),
  \end{aligned}
\end{equation}
where, $\epsilon_y \sim \mathcal{N}(0, \sigma_y)$ models zero-mean i.i.d. Gaussian noise. The set $\bm{\psi} = {[{(\psi_t)}_1, {(\psi_t)}_2,...,{(\psi_t)}_N]}^T \in {\mathbb{R}}^{N \times 1}$  contains values for each of the basis function at time $t$. Given a basis function, one can compute $\bm{\omega}$ for each trajectory $\bm{y}_{1:T}$ using linear regression with a time-dependent design matrix:
\begin{equation}
  \bm{\omega} = ({\bm{\Psi}_{1:T}^T \bm{\Psi}_{1:T})^{-1}} \bm{\Psi}_{1:T} \bm{y}_{1:T},
\end{equation}
where,
\begin{equation}
\bm{\Psi}_{1:T}
=\begin{bmatrix}
  {(\psi_1)}_1 & \cdots  &  {(\psi_1)}_N\\
    \vdots     & \ddots  &  \vdots  \\
  {(\psi_T)}_1 & \cdots\ &  {(\psi_T)}_N\\
\end{bmatrix}
\end{equation}
The $\bm{\omega}$ vector can compactly represent a single trajectory as the number of basis functions is often much lower than the number of trajectory steps\footnote{Handover tasks have an average duration of 4 seconds sampled at 50Hz leading to 200 samples. Instead we use 31 basis functions, thus only using 16\% of the original sample size}. Having a set of motion trajectories, we can compute a probability distribution over the weights $\bm{\omega}$. To capture the variance across trajectories in different demonstrations, we define $\bm{\theta}$ as a parameter that governs the distribution of weight vectors in the set $\bm{\omega}$ and we assume that $\bm{\omega} \sim \mathcal{N}({\bm{\mu}_\omega, \bm{\Sigma}_\omega})$, that is $\bm{\theta} = (\bm{\mu}_\omega, \bm{\Sigma}_\omega)$.

The trajectory distribution $p(\bm{y}_{1:T}; \bm{\theta})$ can now be computed by marginalizing out the weight vector $\bm{\omega}$. The distribution $p(\bm{y}_{1:T}; \bm{\theta})$ defines a Hierarchical Bayesian Model (HBM) whose parameters are given by the observation noise variance $\sigma_y$ and the parameters $\bm{\theta}$ of $p(\bm{\omega}; \bm{\theta})$. We compute the probability distribution of a position at a given time from the $\bm{\omega}$ distribution as:
\begin{equation}
	\begin{aligned}
		p(y_t | \bm{\theta}) &= \int p(y_t | \bm{\omega}) p(\bm{\omega} | \bm{\theta}) d \bm{\omega}\\
		&= \mathcal{N}(y_t | \bm{\psi}_t^T \bm{\mu}_\omega, \bm{\psi}_t^T \bm{\Sigma}_\omega \bm{\psi}_t + \sigma_y).
	\end{aligned}
\end{equation}
The above distribution captures spatial correlations across a set of demonstrations. To cope with demonstrations of varying durations the training set is time aligned.
\subsection{Human and Robot Movement Correlations in an Interaction Model}\label{subsec:IProMP_Ndim}
We now extend ProMPs to a multidimensional setting and compute the correlation for the full set of data-dimensions for human and robot motions across demonstrations. One important assumption in this work is that human-motion collaborative-task trajectories differ spatio-temporally from one another. Under such assumption, the use of motion information is sufficient to distinguish distinct tasks. However, if the assumption is violated and different tasks share similar trajectories, the task recognition system will fail. Our work in \cite{2017Humanoids-Longxin-HRC} addressed this limitation by leveraging differentiating data such as muscle activity electromyography.

IProMPs model the correlation across multiple dimensions. Each dimension is given by each of agent's degrees-of-freedom (DoFs). We define the state vector $\bm{y}_t$ as a concatenation of the observed $P$ number of human DoFs, followed by the $Q$ DoFs of the robot:
\begin{equation}
\bm{y}_t = {[y_{1,t}^H, ... y_{p,t}^H, y_{1,t}^R, ... y_{q,t}^R]}^T,
\end{equation}
where, $(.)^H$ refers to the human position and $(.)^R$ refers to the robot joint angle configuration.
The weight vector $\bm{\omega}$ is the concatenation of all weight vectors involved in a given demonstration. Thus, all the interacting dimensions in a task are correlated as:
\begin{eqnarray}
  \bm{\omega}_i^T = [{(\bm{\omega}_1^H)}^T,...,{(\bm{\omega}_p^H)}^T,
  {(\bm{\omega}_1^R)}^T...,{(\bm{\omega}_q^R)}^T].
\end{eqnarray}
As in the single dimension case, the weight vector is given as a linear regression model: $p(\bm{y}_t|\bm{\omega}) = \mathcal{N}(\bm{y}_t| \bm{H}_t^T \bm{\omega}, 		{\bm{\Sigma}_y})$. $\bm{H}_t$ is the time-dependent basis matrix for the human and robot observations and defined as: 
\begin{equation}
  \bm{H}_t = diag({(\bm{\psi}_t^T)}_1,...,{(\bm{\psi}_t^T)}_{p},{(\bm{\psi}_t^T)}_1,...,{(\bm{\psi}_t^T)}_q).
\end{equation}
Given partial human observations, the posterior distribution is computed using a Kalman Filter (robot observations are set to zero) yielding $\bm{y}_t^o = {[\bm{y}_{1,t}^H, ... \bm{y}_{p,t}^H, \bm{y}_{1,t}^R, ... \bm{y}_{q,t}^R]}^T$. To contrast with a complete observation sequence $[t:t']$, the notation $[t-t'] \in {\mathbb{R}}^{s \times (p+q)}$ is used to indicate a sequence $s$ of partial observations in the interval. Observations can be considered as modulations to via-points. The operation is done by conditioning the ProMPs to reach a certain state $\bm{y}_{t-t'}^o$ at time $(t-t')$. The conditioning adds a desired observation $\bm{x}_{t-t'}=[\bm{y}_{t-t'}^o, \bm{\Sigma}_y^o]$ to the probabilistic model and applies Bayes theorem. Kalman filtering is used to compute the posterior distribution according to:
\begin{equation}
	\begin{aligned}
		\bm{\mu}_\omega^{new} &= \bm{\mu}_\omega + \bm{K} (\bm{y}_{t-t'}^o - \bm{H}_{t-t'} \bm{\mu}_\omega), \\
		\bm{\Sigma}_\omega^{new} &= \bm{\Sigma}_\omega - \bm{K} (\bm{H}_{t-t'} \bm{\Sigma}_\omega).
	\end{aligned}
\end{equation}
Here, $\bm{K}=\bm{\Sigma}_\omega \bm{H}_{t-t'}^T {(\bm{\Sigma}_y^o + \bm{H}_{t-t'} \bm{\Sigma}_\omega \bm{H}_{t-t'}^T)}^{-1}$. Since missing robot observations exist, for each time step of the observation matrix we set $\bm{H}_{t-t'}$ as:
\begin{equation}
  \bm{H}_{t-t'}
  =\begin{bmatrix}
  {(\bm{\psi}_t^T)}_1    & \cdots & 0 & 0 & \cdots & 0	\\
  0 					 & \ddots & 0 & 0 & \ddots & 0					\\
  0 & \cdots & {(\bm{\psi}_t^T)}_{p} & 0 & \cdots & 0\\
  0 & \cdots & 0 & 0_1 & \cdots & 0					\\
  0 & \ddots & \vdots & 0 & \ddots & 0 				\\
  0 & \cdots & 0 & 0 & \cdots & 0_q					\\
\end{bmatrix}
\end{equation}
with $\bm{H}_{t-t'} \in  {\mathbb{R}}^{(p+q)  \times (p+q)N}$.
\subsection{Phase estimation with Dynamic Observation Windows}\label{subsec:phase_est_w_dw}
Humans normally execute repetitions of specific tasks at different speeds. This facts leads to uncertainty in the demonstration's time duration. To capture the task's spatial variation correctly, time alignment must be done. Additionally, the phase of human observations (at test time) must be estimated to align it to that of trained spatial models. 

In Maeda \et original work \cite{2017IJRR-Maeda-PhaseEstimation}, each demonstration (which used static observation windows (SOW))was resampled yielding a nominal duration $T_{nom\_sow}$. We adjust the definition of the nominal duration to fit the length of the dynamic observation window (DOW) duration yielding $T_{nom\_dow}$. For each i$^{th}$ demonstration, we assume constant temporal changes in relation to the nominal duration. Eqtn. \ref{eqtn:scaling_factor} thus introduces a temporal scaling factor to index demonstrations according to the nominal DOW time index. 
\begin{equation}
	\alpha_{i\_dow} = T_i / T_{nom\_dow}.
    \label{eqtn:scaling_factor}
\end{equation}
To determine the best phase estimate during test time, we use the single phase temporal model in \cite{2017IJRR-Maeda-PhaseEstimation}. A distribution of phase ratios across demonstrations is modeled as a normal distribution and set as the phase prior: $\alpha_{dow} \sim \mathcal{N}(\mu_{\alpha_{dow}}, \sigma_{\alpha_{dow}})$. Given a human observation $y_{t-t'}^o$, the phase posterior is:
\begin{equation}
  p(\alpha_{dow} | \bm{y}_{t-t'}^o, \bm{\theta}) \propto
  p(\bm{y}_{t-t'} | \alpha_{dow}, \bm{\theta}) p(\alpha_{dow}),
\end{equation}
with $p(\alpha_{dow})$ as the prior for scaling factor $\alpha_{dow}$, and with a task-specific likelihood given by:
\begin{equation}
	p(\bm{y}_{t-t'} | \alpha_{dow}, \bm{\theta}) =
	\int p(\bm{y}_{t-t'}^o | \bm{\omega}, \alpha_{dow}) p(\omega) d \bm{\omega}.
\end{equation}
So finally, the best task-specific phase estimate given the human observations $y_{t-t'}^o$ is:
\begin{equation}
	\alpha_{dow}^* = \mathop{\arg\max}_{\alpha_{dow}} p(\alpha_{dow} | \bm{y}_{t-t'}^o, \bm{\theta}).
\end{equation}
\begin{figure}[t]
  	\centering
	\includegraphics[width=\linewidth]{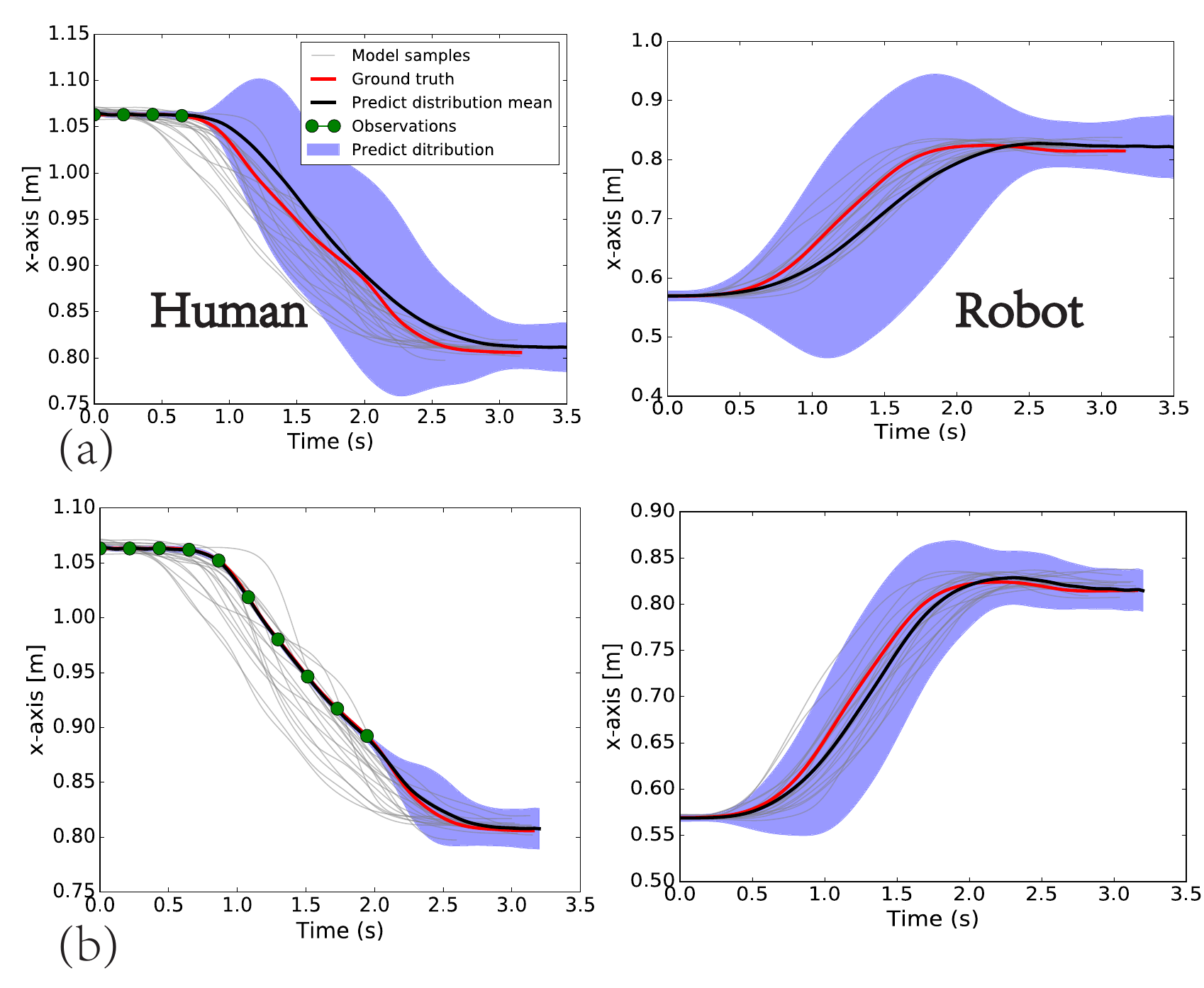}
  \caption{Two different phase estimates for a handover task with the same test data. Top: handovers with 4 human observations up to 0.8 seconds ($dow\_t=0.2$). Bottom: 10 observations up to 2 seconds ($dow\_t=0.2$). Left: Human phase estimation. Right: Robot phase estimation. Note: alpha estimates yield higher accuracy with more observations (upper row errs 0.30s, bottom row errs 0.02s.}
  \label{fig:phase estimation}
\end{figure}
In Sec. \ref{subsec:metric}, we explain how to choose the best duration for the DOW.
\subsection{Task Recognition}\label{subsec:task_recognition}
Once the best phase estimate for human observations is obtained, we must recognize the most likely task model in a set of $k$ tasks. Task recognition also follows a probabilistic approach. We compute the posterior distribution of a task given human observations according to:
\begin{equation}
  p(k | \bm{y}_{t-t'}^o) \propto p(\bm{y}_{t-t'}^o | \bm{\theta}_k, \alpha_{dow}^*) p(k),
  \label{eqtn:task_recognition}
\end{equation}
where, $p(k)$ is the task's prior probability and can be determined by the specific circumstances of an experiment. The likelihood of each component given the model $\theta$ is:
\begin{equation}
	p(\bm{y}_{t-t'}^o; \bm{\theta}_k, \alpha_{dow}^*) =
	\int p(\bm{y}_{t-t'}^o | \bm{H}_{t-t'}^o \bm{\omega}, \bm{\Sigma}_y) p(\bm{\omega}; 	\bm{\theta}_k) d \bm{\omega}.
\end{equation}
A task is selected by choosing the posterior with the highest probability: $k^{*} = \mathop{\arg\max}_{k} p(k | \bm{y}_{t-t'}^o)$.
\begin{figure}[b]
  	\centering
		\includegraphics[width=\linewidth]{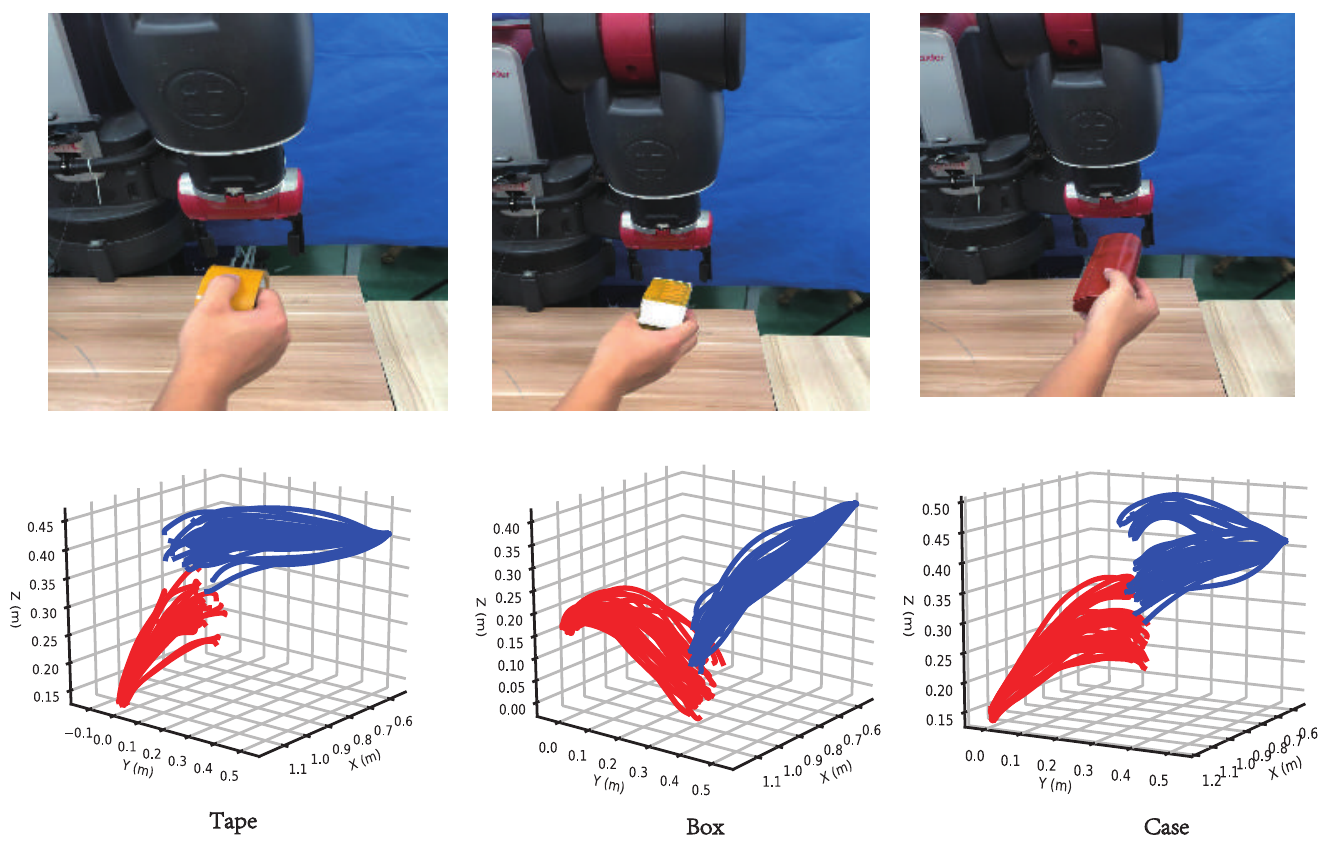}
    	\caption{Three different interaction examples and their respective motions.}
    	\label{fig:demonstrations}
\end{figure}
\section{Iterative Blending of Movement Primitives}\label{sec:trajectory_blending}
In this section, we present the mechanics of the iterative combination of different MPs into a single compounded primitive. It is this iterative blending, afforded by the set of distributions produced as a result of dynamic observation widows that set this work apart. 

Consider a set of $i$ distinct primitives $p_{i}$. When a new human observation is acquired, a new robot primitive $p_{new}$ is generated. Primitive pairs are co-activated through the product of their distributions: $p_{new} \propto {\prod }_{i}p_{i}(\tau)^{a^{[i]}}$, where $a^{[i]} \in[0,1]$ denotes the activation of the $i^{th}$ primitive. The product captures overlapping regions of active MPs.

It's also necessary to modulate the activations of primitives to continuously blend the movement execution from one primitive to the next. First, a trajectory is decomposed into single time steps and uses time-varying activation functions ${a{_{t}}^{[i]}}$ such that product of distributions is defined as:
\begin{equation}
  \begin{aligned}
     p^{*}(\tau) \propto {\prod }_{t} {\prod }_{i} p_{i} (\bm{y}_{t})^{a{_{t}}^{[i]}}, 
     \\
     p_{i}\bm(y_{t}) = \int p_{i}\bm(y_{t}|\bm{\omega}^{[i]})p_{i}(\bm{\omega}^{[i]})d\bm{\omega}^{[i]}.
     \label{eqtn:prod_dists_activation}
   \end{aligned}
\end{equation}
For Gaussian distributions $p_{i}((\bm y_{t} )) = \mathcal{N}(y_t | \bm\mu _{t}^{[i]},\bm\Sigma _{t}^{[i]}) $, the resulting distribution $p^{*}(\bm y_{t})$ is defined as in \cite{2013NIPS-Paraschos_Peters-ProMPs} and is again Gaussian with variance and mean:
\begin{equation}
  \begin{aligned}
    \bm\Sigma_{t}^{*} = \left (  \sum _{i}\left (\Sigma _{t}^{[i]}/a _{t}^{i}\right )^{-1}\right ) ^{-1}, \mbox{ and}
    \\  
    \bm\mu_{t}^{*} = (\bm\Sigma _{t}^{[i]}) \left (\sum_{i} \left ( \bm\Sigma _{t}^{[i]} / a_{t}^{[i]}\right )^{-1}\bm\mu _{t}^{[i]}  \right ).
   \end{aligned}
\end{equation}
As for the activation function used in this work we use a pair of sigmoid functions to define a rising edge and a falling edge respectively:
\begin{equation}
	a_{rise}[t] =  \frac{\mathrm{1} }{\mathrm{1} + e^{-lt} }, \quad 
    a_{fall}[t] =  \frac{\mathrm{1} }{\mathrm{1} + e^{lt} }.
\end{equation}
where, $l$ denotes the gradient of the activation function. 
\subsection{Error Metric}\label{subsec:metric}
In this work we formulate a metric to measure which dynamic observation time-window duration will perform best for in our predictive formulation. Before introducing the metric, we present the performance measurements used in this work and some explanations for how they are measured in the dynamic case and the static case which serves as a baseline. 

To measure the performance of the dynamic time-window formulation we make use of three measurements: (i) $e_p$ Cartesian position final goal accuracy, (ii) $e_q$ Joint angle configuration error sum at each time-step, and (iii) $e_\phi$ Phase estimation error at the goal position. For each of these three measurements we compute the difference between a corresponding ground-truth and the actual goal position.

Furthermore, we are interested in computing the performance gains of the dynamic formulation over that of the static one. For each of the three aforementioned measures we compute the difference between the static and the dynamic results \ie
\begin{equation}
	\begin{array}{ll}
      \nabla e_p 	&= e_{p_{sow(f)}}	-e_{p_{dow\_t}} \\
      \nabla e_q 	&= e_{q_{sow(f)}}	-e_{q_{dow\_t}} \\
      \nabla e_\phi &= e_{\phi_{sow(f)}}-e_{\phi_{dow\_t}}.
	\end{array}
    \label{eqtn:perf_measure_diff}
\end{equation}   
Note that for the static formulation performance is reported according to the observation window percentage (observing an entire task is equivalent to observing 100\% of the task) $sow(f)$. For the dynamic formulation performance is reported according to the duration of the dynamic time window ${dow\_t}$. Thus to measure the performance gain in the dynamic formulation, we compute the difference in performance for the three measures using observation window percentages (static case) and different curves for dynamic window durations. Fig. \ref{fig:exp1_direct_eval} from the Experiments section illustrates these measurements.

In order to the determine which dynamic observation time-window duration has the best performance over the three measures, we use a weighted sum of errors to define the total error measurement for a given dynamic window. The weighted error sum result for each $dow\_t$ is placed in the rows of column matrix $m$ as shown in Eqtn: \ref{eqtn:metric}. Then, the index with the smallest error sum is selected as the best time duration window for a given motion type.
\begin{equation}
	\begin{array}{ll}
    	\min m_{dow\_t} = & \\ \\
      	\gamma_p 		\frac{ e_{p_{dow\_t}} }		{ \max_{t} e_{p_{dow\_t}} }+
      	\gamma_q 		\frac{ e_{q_{dow\_t}} }		{ \max_{t} e_{q_{dow\_t}} }+ 
      	\gamma_{\phi} 	\frac{ e_{\phi_{dow\_t}} }	{ \max_{t} e_{\phi_{dow\_t}} }
    \end{array}
    \label{eqtn:metric}
\end{equation}
where, the weighting scale for each of the three measures is given by $\gamma_i$, where $i=p,q,\phi$. Each factor represents the fraction of error measure for a given $dow\_t$ with respect to the maximum error value across all dynamic window durations.
\section{Experiments and Results}\label{sec:experiments}
Our experimental testbed used the dual-armed humanoid Baxter robot with standard Rethink Robotics electric grippers. An ASUS Xtion Pro camera was used along with OpenNI tracker in ROS Indigo and Linux Ubuntu 14.04. 
As for dynamic observation window durations $dow\_t$, this work tested a range of four discrete values: $dow\_t=[1,0.5,0.2,0.1]$ secs. 
The temporal scaling factor $\alpha\_dow$ is computed by using a nominal duration $T_{nom\_dow}$ set to the duration of the dynamic observation window. Given a dynamic time window duration $dow\_t$, the total number of dynamic observations in a task is then the ratio of task duration to dynamic time window duration: $T/dow\_t$ An equivalent number of co-activations are  needed to update the robot's motion through the task. 
Finally, for the dynamic observations, beyond the duration of the window, the number of observations in a given window must be selected. For these experiments, we observed every 1-out-of-5 human joint angle readings. 

Two distinct experiments are used to test the system performance. The first experiment uses simple trajectories, while the second experiment uses sets of trajectories that share similar starts but have different endings. For each demonstration, a total of 20 trials are executed. Leave-one-out cross-validation (loocv) is used for training and testing. Performance is evaluated under two modalities. First, is a direct evaluation for the dynamic formulation according to the three measures introduced in Sec. \ref{subsec:metric}. Second, is a comparative evaluation between the dynamic formulation and the static one using the same three measures. 
\subsection{Experiment 1}
Exp. 1 is designed to test the accuracy of the dynamic predictions. Three collaborative handover tasks consisting of a simple box, a glasses case, and a circular masking tape are conducted as shown in Fig. \ref{fig:demonstrations}. For each task, the path of the trained human trajectories is unimodal; in other words, all task trials beginning and end paths are the same distribution (in Exp. 2, the ending portion of the trial trajectories diverge in different directions as seen in Fig. \ref{fig:exp2_demos}). 
\\ \\ \textbf{Results}
\subsubsection{Direct Evaluation} 
\begin{figure}[tb]
  \centering
	\includegraphics[width=\linewidth, height=3.25in]{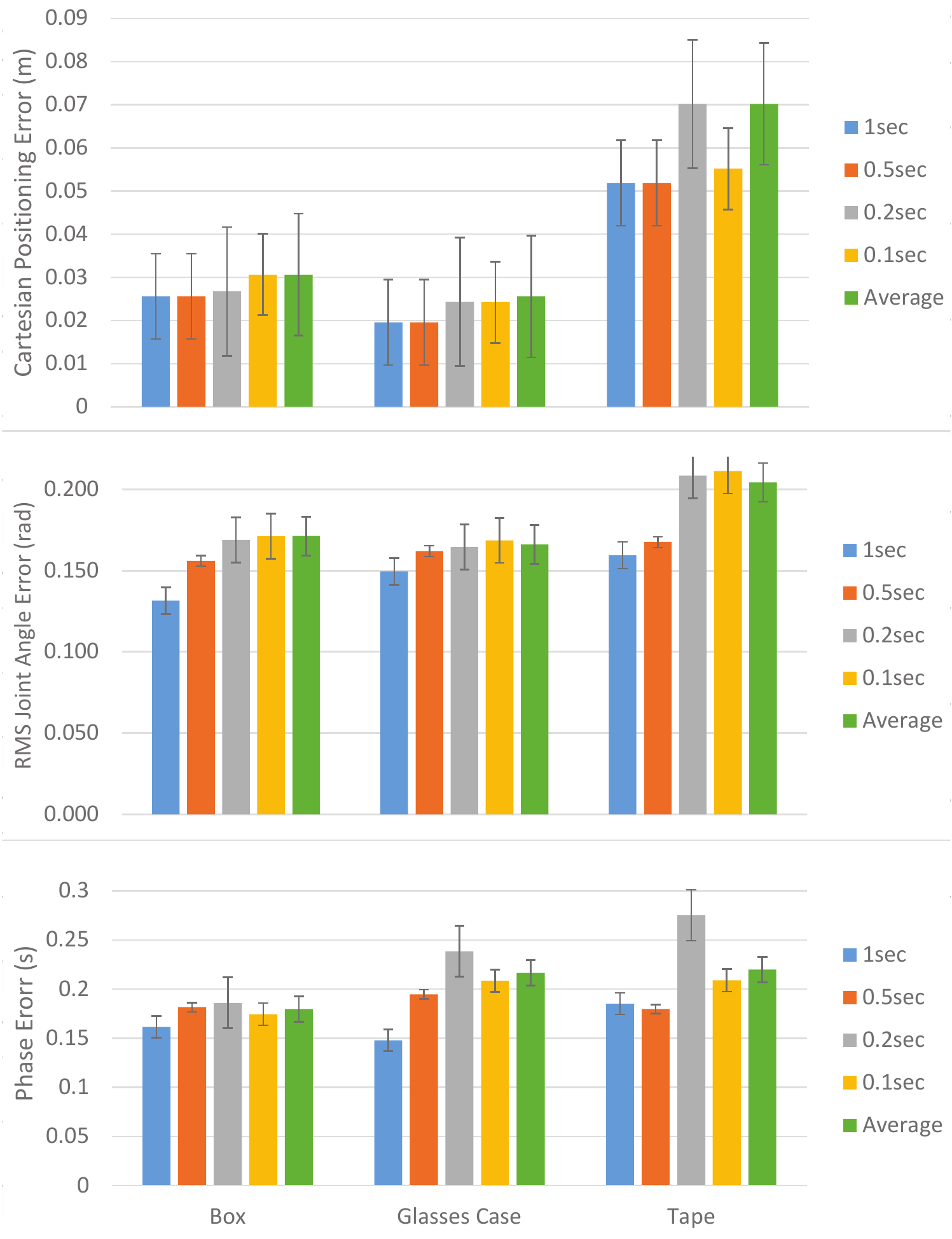}
 	\caption{Cartesian position goal errors, RMS joint trajectory errors, and phase estimation errors are shown for dynamic observation windows of four durations along with the average performance. Data is also shown for the three handover tasks explained in Sec.\ref{sec:experiments}.}
	\label{fig:exp1_direct_eval}
\end{figure}

As summarized in Fig. \ref{fig:exp1_direct_eval}, the dynamic formulation achieved average (of dynamic window durations) Cartesian position goal errors of $(0.028,0.023,0.059)$m; average RMS joint errors of $(0.163,0.163,0.195)$rad, and average phase estimation errors of $(0.176,0.203,0.213)$s for the box, glasses, case, and tap handovers respectively.  

All the best results came from duration $1.0$s (except for the phase error which came from the tape task) but notice that the difference between other window durations is not significant. This give us the flexibility to select shorter durations that increase responsiveness at the expense of a small loss in accuracy.
\subsubsection{Comparative Evaluation} 
\begin{figure}[t]
  \centering
	\includegraphics[width=\linewidth, height=3.5in]{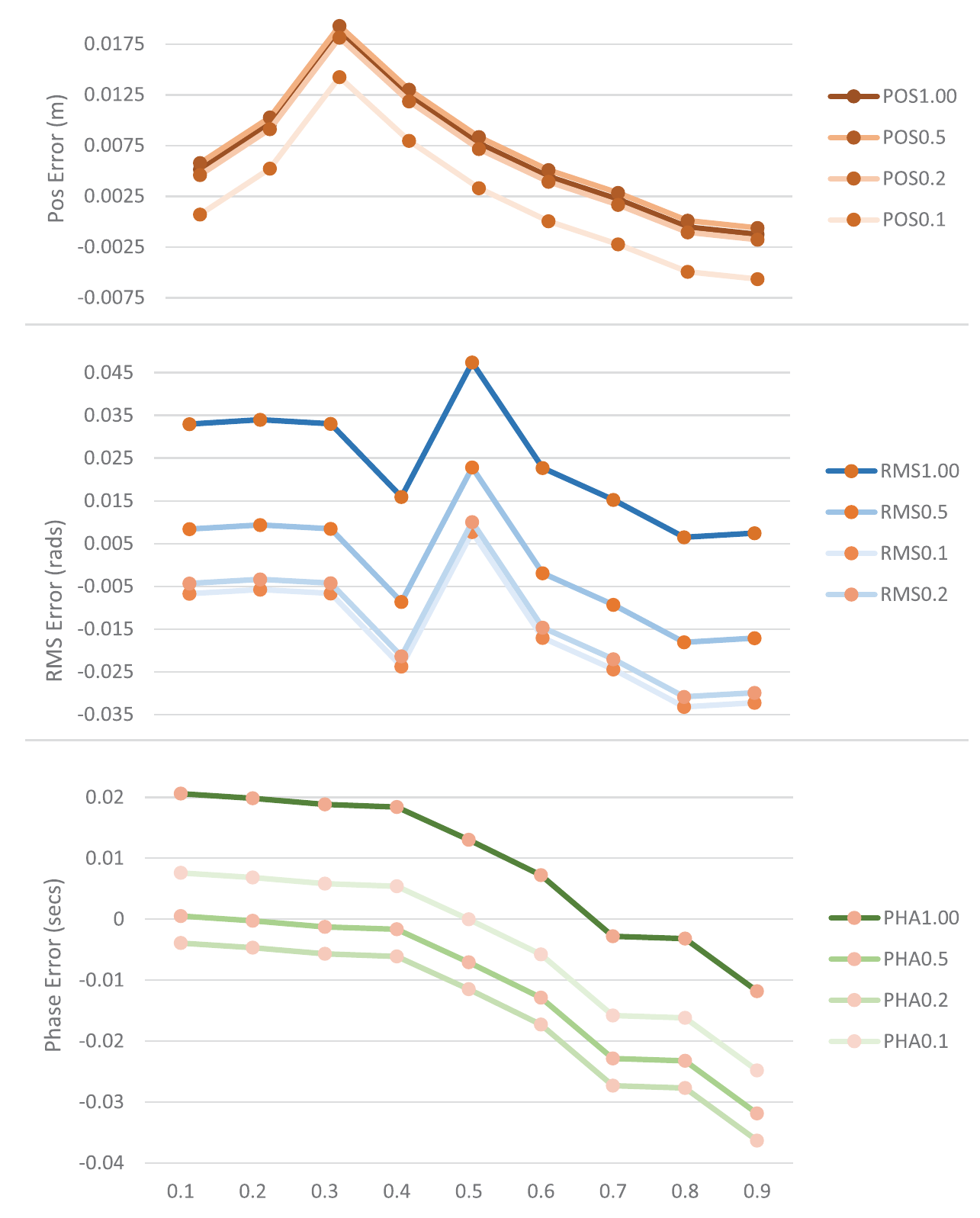}
 	\caption{Box Handover Error Measurements (Static-Dynamic) for Different Static Observation Ratios and Varying Dynamic Window Durations: three error measurement differences between static and dynamic observation window formulations across static observation ratios (0.1 to 0.9): (top) Cartesian goal position error difference (m), (middle) RMS joint angle error difference over time steps, and (bottom) phase estimation error difference.}
	\label{fig:exp1_compare_eval_box}
\end{figure}

Performance comparisons for the dynamic formulation of the handover box between the static and dynamic formulations are summarized in Fig. \ref{fig:exp1_compare_eval_box}. The figure has three plots for each of the three error measure comparisons. With each plot, four curves represent four distinct  dynamic observation window durations. The curves represent the difference between the static and dynamic formulations from Eqtn. \ref{eqtn:perf_measure_diff}. A positive error in the graph implies that the dynamic formulation performance outperforms the static one. 
The top plot, which is related to the Cartesian goal position error difference at the end of the task, shows three of the four curves are almost superimposed. This indicates that the accuracy performance of the dynamic system for durations of 1, 0.5, and 0.2 seconds was similar (ultimately each of these windows will have observed all data). For these three curves it is not until the static system observes around 80\% of the trajectory that it can match the performance of the dynamic system. Given that the dynamic system can continuously update its distributions it gives the ability to increasingly approach the ground-truth while improving responsiveness. 

With reference to the middle plot, we consider the RMS joint angle configuration error difference along trajectory time steps. Notice that the performance of different dynamic window durations varies in non-trivial ways. The best performance comes from the longest duration window $RMS1.00$. For this particular curve, the static formulation does not match the performance of the dynamic window as the error difference does not reach zero by the end of the task. The reasoning is the same as that mentioned for the Cartesian goal position measure.

With respect to the bottom plot, we consider the Phase estimation error difference at the end of the task. Recall that our scaling factor $\alpha_{dow}$ is set by the duration of the dynamic observation window. The best performance comes from the window $PHA1.00$. This indicates that longer durations for nominal window durations assist in having better phase estimation errors. The dynamic performance was only matched by the static formulation when it observed slightly less than 70\% of the trajectory.

\renewcommand{\thefootnote}{\fnsymbol{footnote}}
Given these three independent error measure comparisons, we must determine which dynamic observation window performed best across all measures. Using the metric definition in Eqtn \ref{eqtn:metric}, we tested two $\gamma$ weight combinations \{\{0.33,0.33,0.33\}, \{0.50,0.25,0.25\}\footnote[1]{And all combinations of this set.}\} for our performance metric $m$. In all cases, dynamic observation window with 1.0 second duration had the best performance. 
\renewcommand*{\thefootnote}{\arabic{footnote}}
The results from the box handover case were consistent with the results for the glasses case demonstrations and the tape demonstrations. The dynamic formulation reported similar gains across all measures. 
\subsection{Experiment 2}
\begin{figure}[b]
  \centering
	\includegraphics[width=0.75\linewidth]{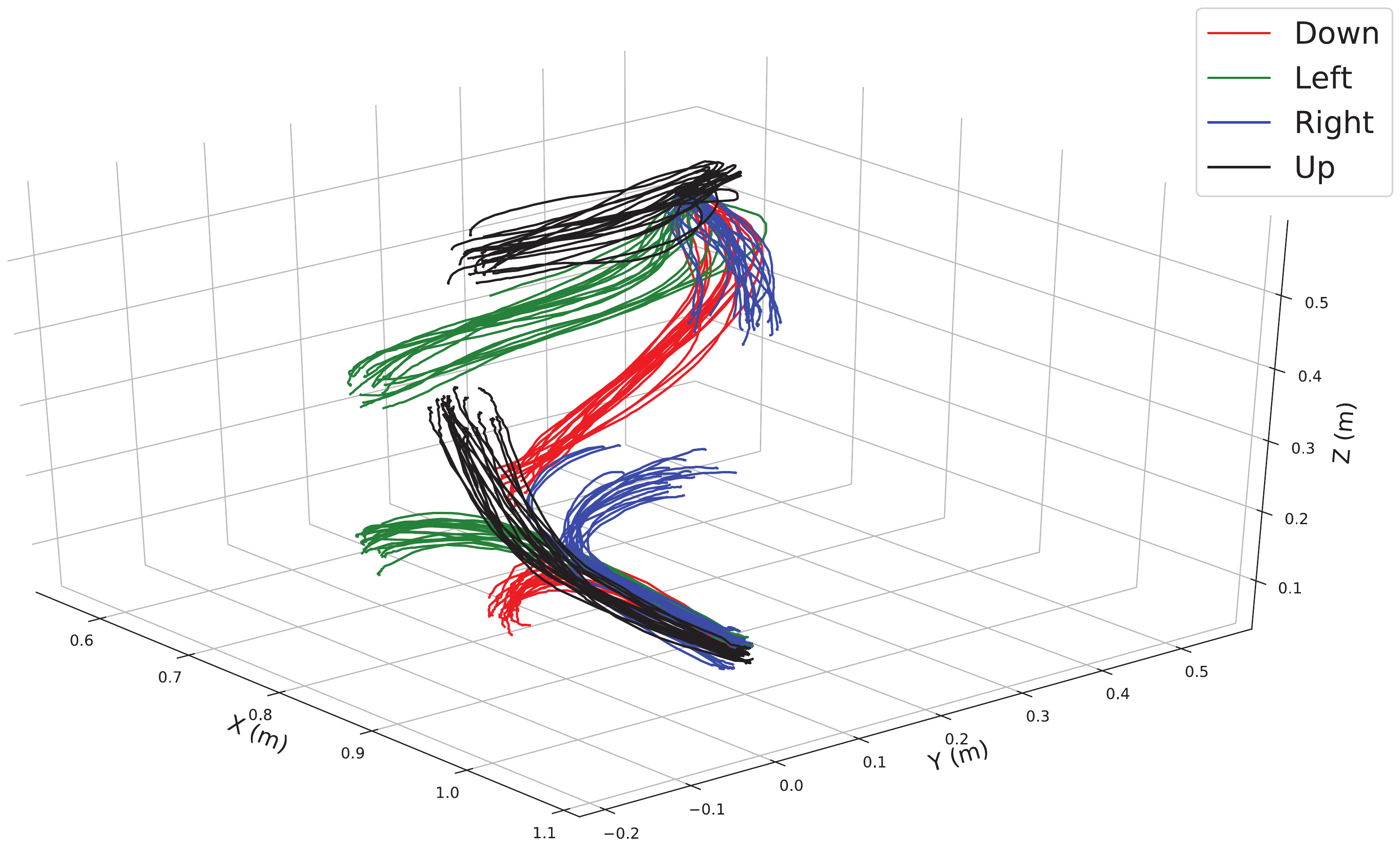}
 	\caption{Collaborative trajectories with similar human initial paths and diverging paths. The red color indicates one collaborative task, the green another one. By using static observations in the first half of the trajectory the predictive system have no better results than chance to predict the correct final position. The static system would also lack responsiveness to the changes effected by the human collaborator later in the task. }
	\label{fig:exp2_demos}
\end{figure}

In Exp. 2 we test the performance of the system, through the use of four handover tasks that share similar initial trajectories but in the end diverge to leftwards, rightwards, upwards, and downwards as depicted in Fig. \ref{fig:exp2_demos}. 
As in Experiment 1, we perform direct evaluations and comparative evaluations. We also compare the performance of dynamic formulation in Exp. 2 with that in Exp. 1.
\\ \\ \textbf{Results}\\
\subsubsection{Direct Evaluation} Table \ref{tbl:exp2_direct_eval}, summarizes performance evaluation in Exp. 2 for the four tasks for the best performing dynamic observation window duration (again in all cases the 1.0s duration:
\begin{table}[h]
	\caption{Performance evaluation for optimal window duration in bi-modal trajectories which diverge right, left, up, and down. Error measures follow definitions presented in Sec. \ref{subsec:metric}.}
	\label{tbl:exp2_direct_eval}
    \begin{tabularx}{\linewidth}{ p{1in}p{0.6in}p{0.6in}p{0.6in} }
      \toprule
      		& POS   & RMS    & Phase   \\ 
      \midrule
      Right & 0.0309 & 0.3467 & 0.1712 \\
      Left  & 0.0248 & 0.1697 & 0.1924 \\
      Up    & 0.0299 & 0.0686 & 0.2301 \\
      Down  & 0.0459 & 0.2654 & 0.1273 \\
      \midrule
      Total & 0.0329 & 0.2126 & 0.1803 
  	\end{tabularx}
\end{table}
If we compare the average performance of Exp. 1 for the three handover tasks, with the four, more complex trajectories of Exp. 2, we learn that Exp. 2 had very similar performance: which is re-assuring as the system performs consistently across trajectories with different path patterns. In terms of final Cartesian position, Exp. 2 had an improved accuracy of 0.007m; in terms of RMS joint trajectory error at each time step, Exp. 2 had slightly lower performance with 0.032rad, and very similar phase performance with a -0.024s difference.
\subsubsection{Comparative Evaluation} 
\begin{figure}[tb]
  \centering
	\includegraphics[width=\linewidth, height=3.5in]{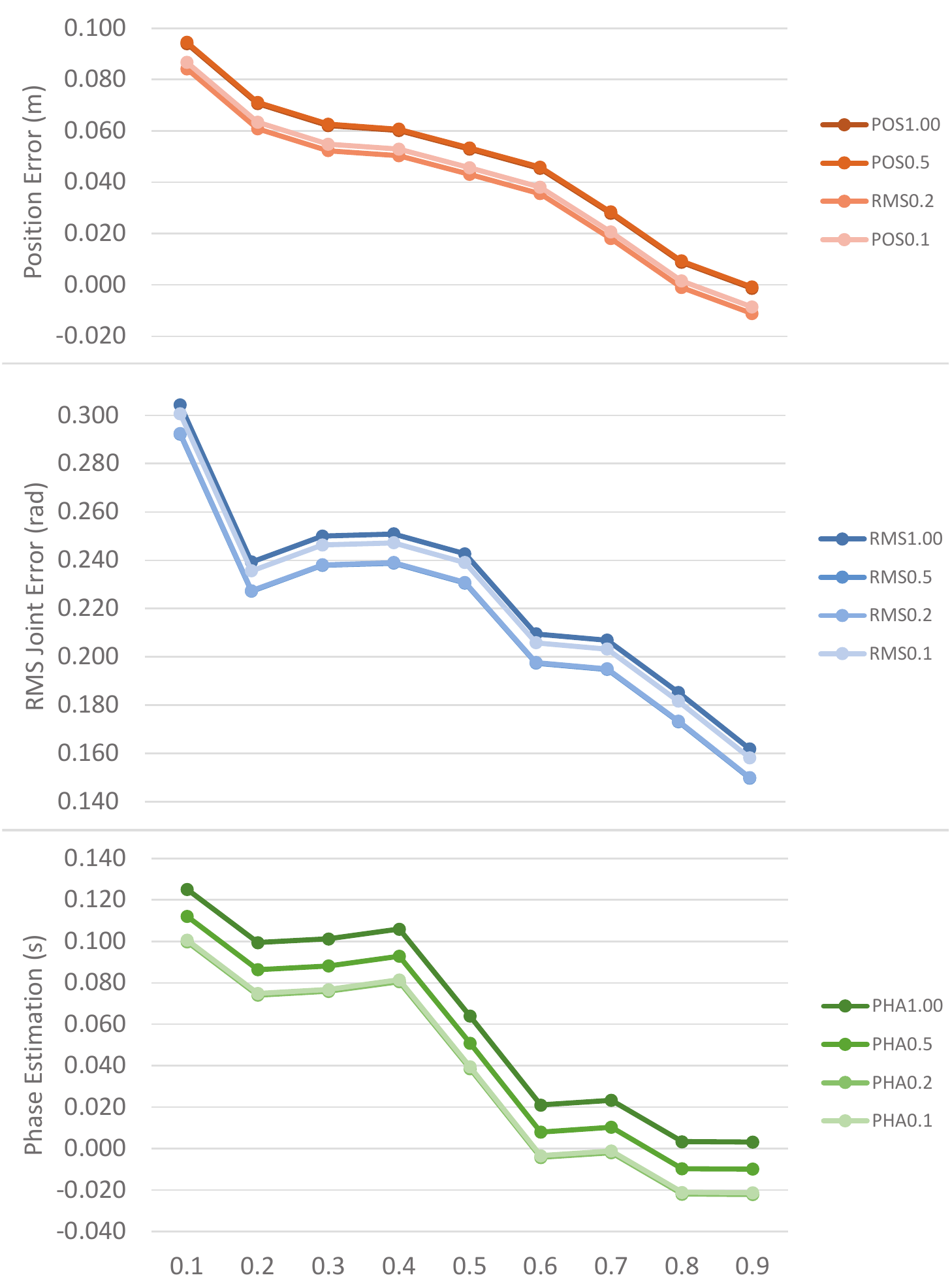}
 	\caption{Comparative performance evaluation for averaged results between the three handover tasks in Exp. 1 and the four diverging tasks in Exp. 2. Averaged results in the static case occurs across observation ratio and for the dynamic case across dynamic observation window duration. The figure's three plots consist of: (top) final Cartesian position error difference, (middle) difference in RMS joint trajectory error at each time step, and (bottom) final position phase estimation error difference. Positive results indicate a better performance by the dynamic formulation. }
	\label{fig:exp2_compar_eval}
\end{figure}

The comparative evaluation between the dynamic and static formulations is more complicated given that we have 3 tasks in Exp. 1 and 4 in Exp. 2 and we are considering the 4 window durations and the observation ratios for the static case. As with our direct evaluation, we resort to presenting average results and commenting in interesting cases. As in Exp. 1, Fig. \ref{fig:exp2_compar_eval}, presents the comparative performance between average static results and average dynamic results. These results are the average of 20 loocv, including situations in which the static formulation both correctly and incorrectly classified the task. In fact, task recognition rates for the static formulation across observation ratios is shown in Table \ref{tble:task_rec_rate}.
\begin{table}[bth]
	\centering
    \caption{Task recognition rate for static observation ratios in Exp. 2.}
    \label{tble:task_rec_rate}
    \begin{tabular}{rcccc}
      \toprule
      & Down & Left & Right & Up      \\ 
      \midrule
      0.1  & 0.53 & 0.58  & 0.47  & 0.89 \\
      0.3  & 0.53 & 0.68  & 0.42  & 0.95 \\
      0.5  & 0.68 & 0.74  & 0.32  & 0.95 \\
      0.7  & 0.74 & 0.89  & 0.63  & 0.95 \\
      0.9  & 0.89 & 1.00  & 1.00  & 1.00 \\ 
    \end{tabular}
\end{table}
Evidently, in many cases the task recognition rate, especially when the observation ratio is around half the trajectory or less, recognition rates are equal to chance or worse. As more observations come in, rates increase. The exception is the ``Up'' trajectory which had better recognition.

There are major trends across all error measures. The dynamic formulation always outperformed the static formulation. Dynamic observations with the longest duration did better, as would be expected, but the main factor is that the performance of the other windows was almost the same making it possible to gain responsiveness for a small loss in accuracy. Finally and most importantly is the fact that the static distribution begins with large errors in the first half of the observations and only later registers quick error drops when it is able to observe more than half of the human trajectory.

As for Cartesian goal position error difference at the end of the task, the static formulation begins with error differences of 0.1m. The average performance of the dynamic formulation was 0.04m: the difference is more than double. Hence, particularly at the beginning, the static formulation struggles to correctly classify the true trajectory and generates motion that has more than double the inaccuracy of the average of the dynamic formulation. 
As for the RMS joint angle error, the same story occurs here. An error difference larger than 0.3rad occurred. Again almost double the average value for the dynamic formulation which yielded 0.14rad. 
As for the phase estimation error, the error difference was 0.13s. The average performance of the dynamic formulation was 0.17s. A negligible difference for most human tasks.
\section{Discussion and Conclusion} \label{sec:conclusion}
This work showed that when the IProMP framework with Phase Estimation is deployed with dynamic human observation windows the accuracy of the predictive generated robot motions increase significantly from the very beginning. In the same way, the dynamic formulation offers a much more significant responsiveness in its anticipatory motions. Hence, the robot generates motion that is constantly adapting smoothly to the human's motion. Our results were tested across a wide variety of motions and objects, including instances where trajectories begin with similar motions and end with diverging ones---not unlike natural human movements. The approach still suffers from appropriate robot motion generation if the observation was not included in training. We are considering the idea of projecting existing probabilistic distributions to newly observed spaces if we are confident the task type is similar. Also, sometimes the predicted motion reaches the limits of a robot's arm workspace. In this case, we are considering querying the other robot arm to successfully complete the task.
\section{Acknowledgements} \label{sec:Acknowledgements}
This work is supported by ``Major Project of the Guangdong Province Department for Science and Technology (2014B090919002), (2016B0911006) and by the National Science
Foundation of China [grant number 61750110521].
\bibliographystyle{IEEEtran}
\bibliography{IEEEabrv,Xbib}
\end{document}